\documentclass[cameraready]{Interspeech}
\usepackage{ragged2e} 
\usepackage{booktabs}
\usepackage[inline]{enumitem}
\usepackage{tabularx}
\usepackage{url}
\usepackage{makecell}

\definecolor{headergray}{gray}{0.85}  
\definecolor{rowgray}{gray}{0.95}
\definecolor{headerblue}{RGB}{0,112,192} 
\usepackage[table]{xcolor}   

\title{Simple Language Normalization Wins: Cross-Lingual Speaker Verification for the TidyVoice 2026 Challenge}

\author[affiliation={1}, orcid=https:0000-0002-0821-9125]{Nina}{Hosseini-Kivanani}

\address{
    $^1$ University of Luxembourg \& Radio Télévisioun Lëtzebuerg (RTL), Luxembourg
}

\email{nina.hosseinikivanani@ext.uni.lu}


\usepackage[table]{xcolor}
\usepackage{comment}
\usepackage{amsfonts}
\usepackage{amsmath}
\usepackage{graphicx}
\usepackage{booktabs}
\usepackage{multirow}
\usepackage{makecell}

\usepackage{enumitem}

\begin{document}

\maketitle

\begin{abstract}
Cross-lingual mismatch remains a key source of overall degradation in modern speaker verification. The TidyVoice2026 Challenge targets this setting with text-independent verification, comprising 3,666 training and 808 development speakers in 40 languages and 2,200 evaluation speakers in 38 unseen languages, without language labels at test time. Starting from the official SimAM-ResNet34 baseline pretrained on VoxBlink2 and VoxCeleb2 and fine-tuned on TidyVoice, we revisit Nuisance Attribute Projection (NAP) as a simple language-normalization step in the embedding space. We estimate a compact language subspace from cross-language same-speaker differences and project embeddings onto its orthogonal complement before cosine scoring with Adaptive Symmetric score normalization. This reduces development EER from 2.97\% with cosine and 2.70\% with AS-Norm to 2.18\% and yields a Codabench evaluation score of 8.40, showing that simple back-end language normalization can rival more complex systems.
\end{abstract}

\section{Introduction}

Deep speaker verification systems based on x-vectors, ECAPA-TDNN, and attention-enhanced ResNets achieve very low error rates on benchmarks such as VoxCeleb when train and test conditions are matched~\cite{snyder2018x,desplanques2020ecapa,qin2022simple}, but performance remains fragile under cross-lingual mismatch, where enrollment and test utterances of the same speaker are produced in different languages~\cite{misra2014spoken,misra2018modelling}. In such scenarios, language-specific phonotactics and prosodic patterns act as nuisance factors that can dominate the representation space and degrade speaker discrimination.

Several strategies have been proposed to mitigate language or domain variability. Classical i-vector and PLDA systems use low rank subspace modeling and Nuisance Attribute Projection (NAP) to estimate and remove subspaces associated with channel or language~\cite{dehak2010front}, while deep embedding approaches extend this idea through semi-supervised nuisance attribute networks, supervised domain adaptation~\cite{lin2019semi,sarfjoo2020supervised}, and adversarial objectives that encourage language invariant representations in cross-lingual settings~\cite{xia2019cross,rohdin2019speaker}. In parallel, self-supervised speech encoders such as WavLM have emerged as strong universal front ends for automatic speaker verification on the SUPERB benchmark~\cite{chen2022wavlm,yang2021superb}, but most evaluations involve a small number of training and test languages or assume that language labels are available during adaptation or scoring.

The TidyVoice2026 Challenge targets language robustness under stronger conditions. The TidyVoice corpus~\cite{farhadipour2026tidyvoice}
comprises 3{,}666 training speakers and 808 development speakers
across 40 languages, together with a held-out evaluation set of
2{,}200 speakers in 40 previously unseen languages, for a
text-independent speaker verification task, where the primary metric is Equal Error Rate (EER), and the secondary metric is the minimum
Detection Cost Function (minDCF). Crucially, language labels are not available at test time, which emphasizes language invariant embeddings and back-end compensation methods that do not rely on explicit language identification.

The official baseline is a SimAM ResNet34 embedding extractor operating on 80-dimensional filterbank features with cepstral mean and variance normalization~\cite{qin2022simple}. The model is pre-trained on VoxBlink2 and VoxCeleb2, then fine-tuned on TidyVoice. 
With cosine scoring on the development set, this baseline reaches 2.97\% EER and 0.82 minDCF, and applying Adaptive Score Normalization (AS-Norm)~\cite{matvejka2017analysis} improves performance to 2.70\% EER and 0.64 minDCF. There is still clear room for improvement, especially for cross-language trials.

We describe our submission to the TidyVoice2026 Challenge. Our main finding is that classical nuisance compensation remains effective when applied carefully to a strong neural baseline. We estimate a low-dimensional language subspace in the embedding space and apply Nuisance Attribute Projection (NAP) to remove leading language directions before scoring, following the spirit of earlier i-vector-based systems~\cite{dehak2010front}. Combined with AS-Norm, this post hoc pipeline reduces development EER from 2.97\% to about 2.18\%, and further to around 2.0\% in an optimized fusion setup. On the official Codabench evaluation, our best single system, using NAP with $k{=}30$ projected language directions and AS-Norm, obtains an evaluation score of 8.40 and ranks at the top of the leaderboard.

We also investigate complementary strategies. First, we explore language adversarial training using a Domain Adversarial Neural Network (DANN) objective that encourages embeddings to be informative about speaker identity but uninformative about language, building on earlier adversarial language adaptation work~\cite{rohdin2019speaker}. Second, we examine self-supervised WavLM-based front ends, both as frozen feature extractors and combined with ECAPA TDNN back-ends~\cite{chen2022wavlm,desplanques2020ecapa}. Finally, we study score and embedding level fusion across systems. These approaches yield consistent gains over the raw baseline but do not surpass the simplicity and robustness of the NAP-based language compensation pipeline in our experiments.

Our contributions are threefold.~\begin{enumerate*}[label=(\roman*)]
\item We analyze language-induced variability in the TidyVoice setting and show that a low-dimensional language subspace can be estimated reliably from training data.
\item We demonstrate that NAP combined with AS-Norm yields stable improvements over a competitive SimAM ResNet34 baseline in a 40-language training and 38-language evaluation scenario without language labels at test time.
\item We release a complete and vectorized preprocessing, scoring, and submission pipeline that can process about 12 million trials efficiently and that can serve as a reference implementation for future cross-lingual speaker verification work.
\end{enumerate*}

Compared to prior work on NAP and cross-lingual speaker verification, our focus is not on proposing a new architecture but on showing that a classical nuisance subspace method, applied as a post hoc transform on a neural baseline, remains competitive under the TidyVoice2026 conditions with 40 training languages, 40 unseen evaluation languages, and no language labels at test time. The back-end is simple to implement, does not require retraining the encoder, and can be attached to any embedding extractor that outputs fixed-dimensional speaker vectors.
\section{Methodology}

\subsection{Task and data}

The TidyVoice2026 task~\cite{farhadipour2026tidyvoice} is text-independent speaker verification under cross-lingual mismatch. The TidyVoice corpus contains 3,666 training and 808 development speakers across 40 languages, and an evaluation set with 2,200 speakers in 38 unseen languages. Audio is sampled at 16kHz. The development set comprises 59,443 utterances and 12 million verification trials; evaluation labels remain hidden. Systems output similarity scores for all trial pairs in two tracks: semi-supervised (limited labels) and unsupervised. The primary metric is Equal Error Rate (EER), with minimum Detection Cost Function (minDCF, $p_{\text{target}}=0.01$) as a secondary metric.

\subsection{Baseline and alternative encoders}

Our systems are built on the SimAM ResNet34 champion with Attentive Statistics Pooling (ASP)~\cite{qin2022simple}. The front end extracts 80-dimensional log Mel filterbank features from 25ms windows with 10ms shift. A ResNet 34 encoder with SimAM attention produces frame-level activations, ASP aggregates them to a single utterance level vector, and a fully connected layer maps to a 256-dimensional speaker embedding. The champion is pre-trained on VoxBlink2~\cite{lin2024voxblink2} and VoxCeleb2~\cite{chung2018voxceleb2}, then fine-tuned on TidyVoice with SGD and ArcFace loss (scale $s=32$). Training uses 6s crops, batch size 24, and a learning rate. Data augmentation combines MUSAN noise and simulated RIR reverberation. In inference, we run the encoder on the full waveform and extract a 256-dimensional embedding per utterance. With cosine scoring on the development set, this yields 2.97\% EER and 0.82 minDCF, which improves to 2.70\% EER and 0.64 minDCF with Adaptive Score Normalization (AS-Norm).

For comparison, we train two ECAPA TDNN~\cite{desplanques2020ecapa} encoders on TidyVoice, with channel sizes 512 and 1024, Res2Net style blocks with Squeeze and Excitation, ASP, and a 192-dimensional embedding layer. Both models use 2s crops, SpecAugment on FBanks, SGD, and ArcFace. The two configurations differ in channel size and batch size (64 vs. 32) but share the same schedule. Both reach 4.82\% EER with cosine scoring; AS-Norm reduces the c1024 model to 4.22\% EER.

To explore language invariance, we attach a domain adversarial branch to the champion encoder~\cite{ganin2016domain}. The 256-dimensional embedding feeds a speaker classifier (ArcFace over 3,666 speakers) and a language classifier over 40 languages. The language branch uses a Gradient Reversal Layer followed by two 512-unit layers with batch normalization, ReLU, and dropout, and a softmax. The total loss combines the speaker and language classification losses with a fixed weight $\alpha = 0.5$, and we use the standard Ganin schedule $\lambda(p) = \frac{2}{1+\exp(-10p)} - 1$ where $p$ increases linearly from 0 to 1 over training epochs. Fine-tuning this model for 10 epochs on TidyVoice yields 4.39\% EER and 0.84 minDCF, with improved cross-lingual balance but higher overall error.

Finally, we evaluate a self-supervised WavLM front-end~\cite{chen2022wavlm} based on ``wavlm base plus", a 12-layer transformer. Three-second segments are passed to WavLM, a learned weighted sum over internal layers feeds ASP, and a linear projection to 256 dimensions, and we train the head with ArcFace and AdamW. Training uses AdamW with cosine annealing, 10--30 epochs, batch size 16--32, and learning rates from $5{\times}10^{-5}$ to $5{\times}10^{-4}$ depending on the regime. We consider three regimes. frozen baseline, frozen but optimized training, and partial fine-tuning of the top three transformer layers. EER ranges from 28.54\% down to 16.27\% and 14.69\% after AS-Norm~\cite{matvejka2017analysis}, still far behind the supervised ResNet and ECAPA systems. We therefore use WavLM models only for analysis, not as primary submissions.

\subsection{Language compensation and back-end scoring}
\label{sec:backend}

Our main contribution is a simple language normalization based on Nuisance Attribute Projection (NAP)~\cite{solomonoff2007nuisance}. We estimate a low-dimensional language subspace from development embeddings, then project embeddings onto its orthogonal complement before scoring. We group development embeddings by speaker and language, and keep speakers who appear in at least two languages. For each such speaker $i$ and language pair $(l_1,l_2)$ we compute mean embeddings $\bar{e}_{i,l}$ and form difference vectors
\begin{equation}
\delta_{i,l_1,l_2} = \bar{e}_{i,l_1} - \bar{e}_{i,l_2}.
\end{equation}
After L2 normalization, all difference vectors are stacked into a matrix $\Delta \in \mathbb{R}^{N \times d}$, with $d = 256$ the embedding dimension. We compute the empirical covariance $C$ from $\Delta$ and take the first $k$ eigenvectors as columns of $V \in \mathbb{R}^{d \times k}$. These span an estimated language subspace. 
The NAP projection matrix
\begin{equation}
P = I_d - V V^{\top}
\end{equation}
is applied to each embedding $e$ to obtain a language compensated vector $\tilde{e} = P e$, followed by L2 normalization. A sweep over $k \in \{1,2,3,5,8,10,15,20,30,40,50\}$ on development trials with AS-Norm shows monotonic improvements up to $k=30$ (2.18\% EER), then slight degradation at $k=50$. We fix $k=30$ for all NAP experiments. Estimating $P$ requires a single eigendecomposition of a $256\times 256$ covariance matrix and adds negligible overhead at inference, since it reduces to one matrix–vector multiplication per embedding.

For scoring, we use cosine similarity between enrollment and test embeddings, followed by Adaptive Symmetric score normalization. The AS-Norm cohort comprises 500 randomly selected training speakers across all 40 languages, with approximately 12 utterances per speaker. Given this language-balanced cohort, we compute Z-norm and T-norm statistics over the top $N$ cohort scores on the enrollment and test side and average the normalized scores. We use $N=300$ for the plain champion and $N=400$ for the NAP pipeline. No development or evaluation labels are used in cohort construction. We also evaluate a classical back-end consisting of global mean centering on training embeddings, NAP with $k=30$, Linear Discriminant Analysis (LDA) trained with speaker labels, L2 normalization, cosine scoring, and AS-Norm. This configuration allows us to test whether adding supervised dimensionality reduction on top of NAP improves generalization.

\subsection{Submitted systems}

All official submissions build on SimAM ResNet34 embeddings and differ only in back-end processing. Our primary system applies NAP with $k=30$ followed by AS-Norm, a champion plus AS-Norm variant without NAP. and a classical pipeline with centering, NAP, LDA, L2 normalization, and AS-Norm. Two supplementary runs explore embedding fusion with PLDA and champion with PLDA scoring, which perform clearly worse and are included only for completeness.

\section{Results}

\subsection{Development set performance}

\begin{figure}[t]
\centering
\includegraphics[width=\linewidth]{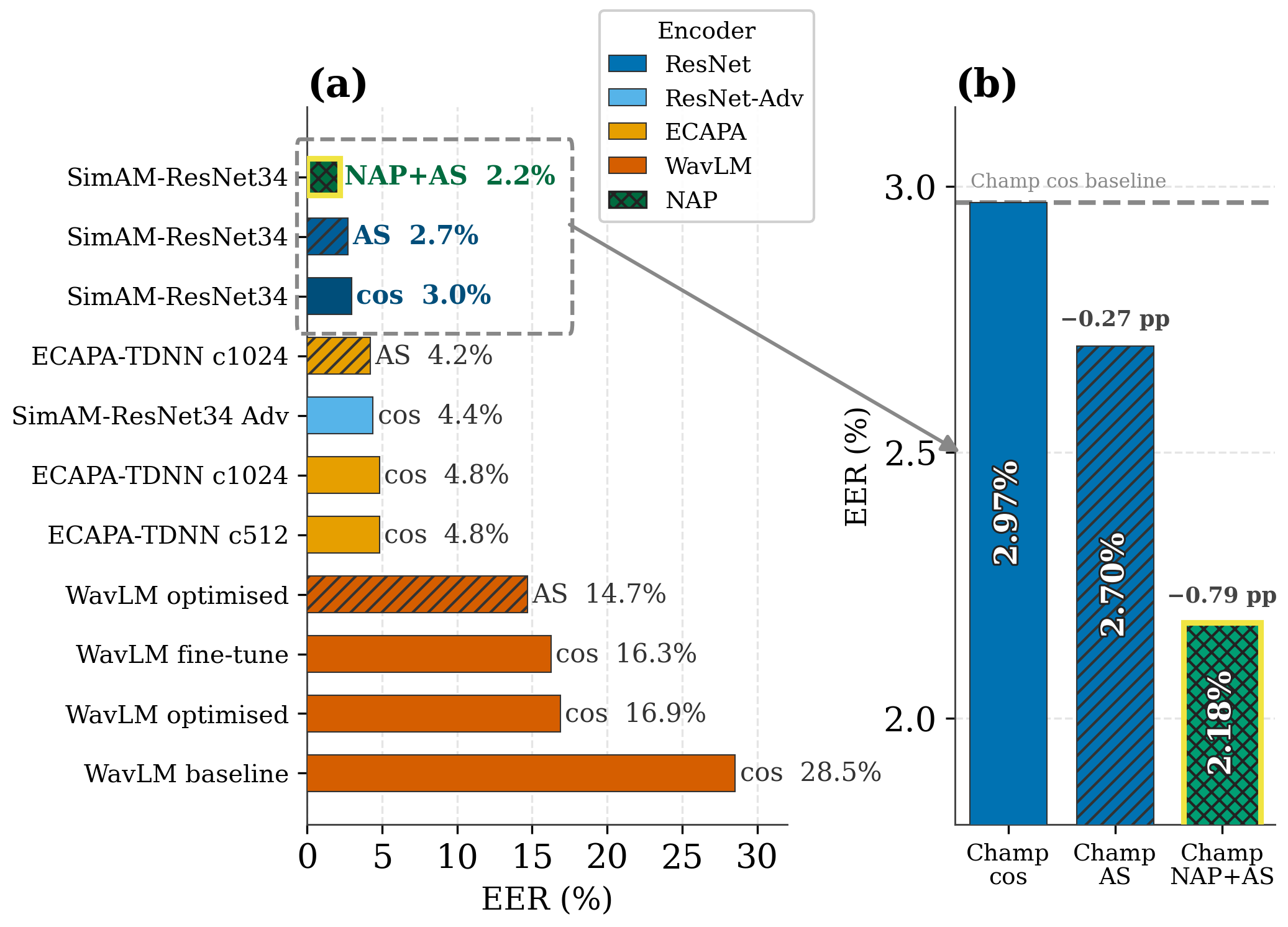}
\caption{Development set results. (a)~EER~(\%) for all systems, sorted best (top) to worst (bottom). Colors denote encoder family, and hatching indicates the scoring back-end (cosine, AS-Norm, or NAP+AS). (b)~Champion system (SimAM-ResNet34) EER under three scoring configurations; annotations show absolute improvement over cosine scoring.}
\label{fig:eer_all}
\end{figure}

Figure~\ref{fig:eer_all} summarizes results on the TidyVoice development set, which contains 12 million trials from 808 speakers and 40 languages. The best single system without explicit language compensation is the champion with AS-Norm at 2.70\% EER and 0.64 minDCF. Applying NAP with $k{=}30$ to champion embeddings and re-running AS-Norm
reduces EER to 2.18\% with a minDCF of 0.73, our best development performance.
Relative to the champion with AS-Norm, this corresponds to a 19\% relative
EER reduction on the development set. ECAPA TDNN models trained from scratch are clearly weaker, around 4.8\% EER with cosine and 4.2\% with AS-Norm, and WavLM systems remain in the 14.7 to 28.5\% range. AS-Norm consistently improves performance, with the largest absolute gains for weaker models.

\subsection{Cross lingual behavior}
\label{sec:crosslingual}

\begin{figure*}[hpt!]
\centering
\includegraphics[width=1.4\columnwidth]{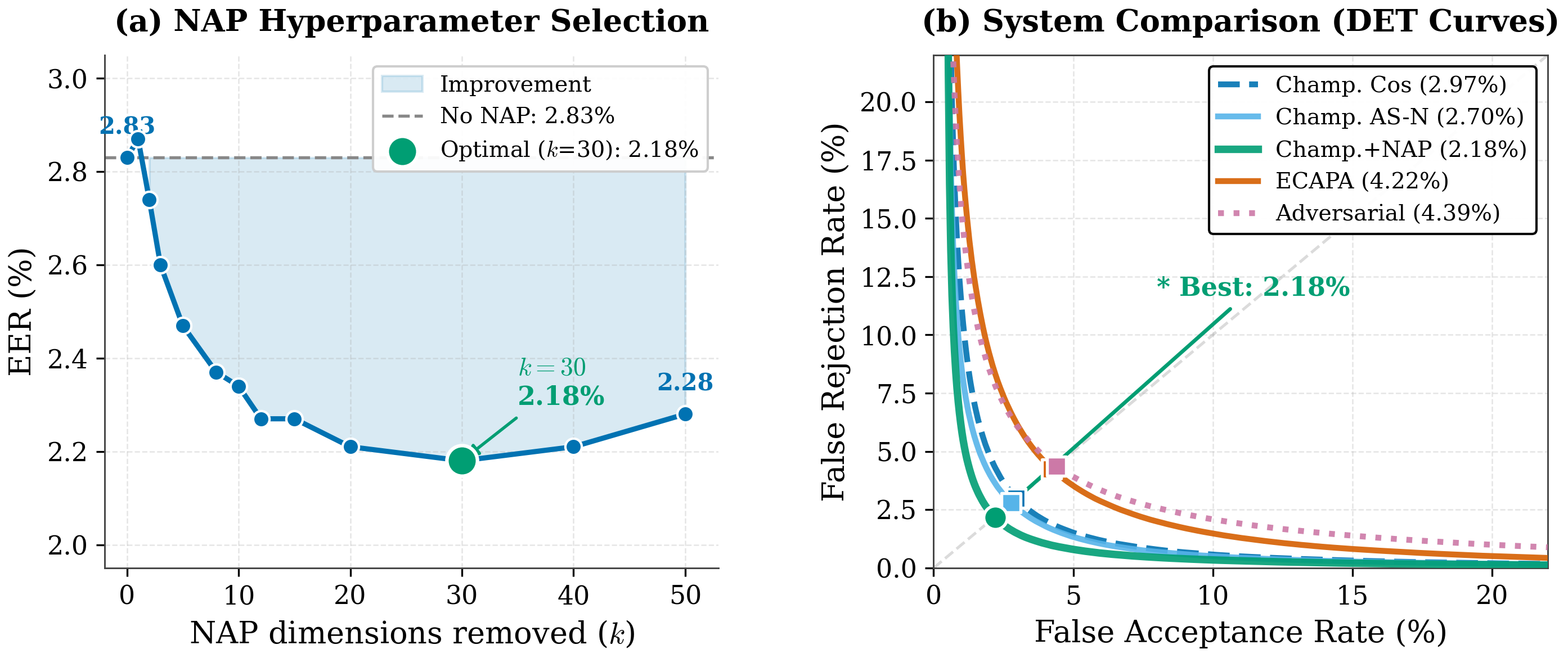}
\caption{NAP sweep and DET curves for the champion encoder.}
\label{fig:combined}
\end{figure*}

To analyze language effects, we decompose development trials into four conditions depending on whether target and non-target pairs share the same language. Table~\ref{tab:eval_all} and 
report EER for each combination.
Figure~\ref{fig:combined} shows the development EER as a function of removed language dimensions. Performance improves monotonically up to $k=30$, then plateaus and slightly degrades at $k=50$, indicating that higher-order components increasingly overlap with speaker-discriminative information.

\begin{table}[hpt!]
    \centering
    \caption{Codabench evaluation scores for all submitted systems (top, lower is better) and official per-condition results for the best system NAP ($k{=}30$) + AS-Norm (bottom, EER / minDCF).}
    \label{tab:eval_all}
    \scriptsize
    \setlength{\tabcolsep}{4pt}

    \rowcolors{2}{rowgray}{white}
    \begin{tabular}{l c}
        \rowcolor{headergray}
        \textcolor{black}{\textbf{System}} &
        \textcolor{black}{\textbf{Eval score}} \\
        \toprule
        NAP ($k{=}30$) + AS-Norm        & \textbf{8.40} \\
        Champion + AS-Norm              & 8.92 \\
        Centering + NAP + LDA + AS-Norm & 9.73 \\
        Embedding fusion + PLDA         & 12.09 \\
        Champion + PLDA                 & 14.44 \\
        \bottomrule
    \end{tabular}

    \vspace{0.4em}

    {\footnotesize\textit{Breakdown for best system: NAP ($k{=}30$) + AS-Norm}}\\[0.2em]

    \rowcolors{2}{rowgray}{white}
    \begin{tabular}{l c c}
        \rowcolor{headergray}
        \textcolor{black}{\textbf{Condition}} &
        \textcolor{black}{\textbf{Task 1. EER / minDCF}} &
        \textcolor{black}{\textbf{Task 2. EER / minDCF}} \\
        \toprule
        Tgt(diff) / NT(diff) & 7.08 / 0.4445 & 8.44  / 0.5054 \\
        Tgt(diff) / NT(same) & 8.93 / 0.6894 & 12.89 / 0.6395 \\
        Tgt(same) / NT(diff) & 6.77 / 0.4406 & 9.23  / 0.5108 \\
        Tgt(same) / NT(same) & 8.63 / 0.6908 & 13.33 / 0.6733 \\
        \midrule
        Overall              & \textbf{8.40 / 0.6485} & 12.15 / 0.6299 \\
        \bottomrule
    \end{tabular}
\end{table}

For all encoders, the hardest condition is \textit{Tgt(diff) / NT(same)}, where target speakers switch language while impostors stay in one language. For the champion, the EER ranges from 0.75\% in the easiest case to 5.18\% in the hardest, showing that cross-lingual targets against same-language non-targets are particularly challenging. The adversarial model reduces the relative gap between easiest and hardest conditions, but at the cost of higher overall error, indicating a trade-off between language invariance and discriminability. Applying NAP with $k=30$ yields consistent improvements across all four conditions, with the largest absolute gains observed in the challenging Tgt(diff) / NT(same) regime.

\subsection{Evaluation results}

Table~\ref{tab:eval_all} reports Codabench scores on the held-out evaluation set. Lower is better. The ranking matches the development findings. NAP plus AS-Norm is best, followed by Champion plus AS-Norm, and then the full LDA pipeline. PLDA-based back-ends perform the worst. For the best system, NAP with $k=30$ plus AS-Norm, the organizers also provide a per-condition breakdown on the evaluation set, summarized in Table~\ref{tab:eval_all}. Task 1 is semi-supervised with 4M trial pairs, and Task 2 is unsupervised with 1.28M pairs.

Compared to 2.18\% EER on development for the same configuration, the overall
evaluation EER of 8.40\% is around six percentage points higher across all
conditions. The hardest case remains Tgt(diff) / NT(same), and same-language
non-target conditions show the highest minDCF. The consistent ordering of
conditions between development and evaluation suggests that the language-mismatch
patterns, rather than overfitting to development speakers, drive this gap.
Task 2 uniformly increases EER but preserves the condition ranking and has
similar minDCF.


The TidyVoice development set covers 40 languages with highly imbalanced counts. English contributes about 23\% of utterances, followed by German, Belarusian, Catalan, and French, while several languages such as Hausa, Yoruba, and Upper Sorbian have fewer than 100 utterances. This skew implies that overall EER is dominated by high-resource languages, while evaluation still tests robustness on severely underrepresented languages.

\section{Discussion}

Our results show that a simple NAP-based back-end on top of a strong pre-trained SimAM ResNet34 baseline is highly effective in the TidyVoice2026 setting. Projecting out a low-dimensional language subspace estimated from cross-language same-speaker differences reduces development EER from 2.97\% with cosine and 2.70\% with AS-Norm to 2.18\%. This is consistent with earlier work on NAP and front-end factor analysis, where low-rank nuisance subspaces capture channel or language variation~\cite{dehak2010front}. The Codabench evaluation scores confirm that this benefit transfers to unseen languages. The NAP plus AS-Norm system achieves the best overall score of 8.40, ahead of the same champion without NAP and of more complex LDA and PLDA-based back-ends.

The per-category analysis underlines that language remains a dominant nuisance factor in cross-lingual speaker verification~\cite{misra2014spoken,misra2018modelling}. For the champion model, the hardest condition is Tgt(diff) and NT(same), where target speakers switch language while impostors stay in one language. NAP reduces EER in this regime on the development set and yields similar relative patterns on the evaluation breakdown, but the absolute gap between easiest and hardest conditions remains. This echoes prior studies that report strong language dependence even when embeddings are trained on multilingual data~\cite{misra2018modelling}. Our results indicate that even when encoders are trained on multilingual corpora, explicit nuisance-subspace removal can still uncover residual language structure that harms cross-lingual trials.

In contrast, more sophisticated adaptation strategies perform worse in this benchmark. The adversarial DANN variant improves the balance between language conditions, but increases overall EER to 4.39\%. This suggests a trade-off between language invariance and speaker discriminability that is hard to optimize under limited task-specific data, consistent with observations in adversarial language adaptation for speaker verification~\cite{xia2019cross,rohdin2019speaker}. The ECAPA TDNN encoders trained only on TidyVoice lag behind the pretrained SimAM ResNet34 baseline, which supports the importance of large-scale out-of-domain pretraining on corpora such as VoxCeleb and VoxBlink2 before task-specific fine-tuning~\cite {snyder2018x,desplanques2020ecapa,lin2019semi}.

Taken together, these results suggest that, in large-scale cross-lingual benchmarks such as TidyVoice2026, careful back-end language normalization can offer a more robust and cost-effective improvement path than designing new, heavily tuned encoders. Even when other systems obtain slightly better absolute scores on the leaderboard, our analysis shows that a simple NAP + AS-Norm back-end delivers strong performance with minimal engineering effort and full transparency of its effect on the embedding space.

Self-supervised WavLM-based systems perform markedly worse than the supervised encoders in this challenge, with EER between about 15\% and 29\% even after AS-Norm. This contrasts with their strong performance on generic ASV benchmarks such as SUPERB~\cite{chen2022wavlm,yang2021superb}. A likely explanation is that our heads are trained only on relatively shallow crops of TidyVoice, without the more extensive adaptation pipelines used in prior work. Jointly optimizing WavLM and task-specific discriminative losses on larger and more balanced multilingual data may be necessary to realize the full potential of self-supervised encoders under strong cross-language mismatch.

Finally, the TidyVoice language distribution highlights an important limitation. Overall development EER is dominated by high-resource languages such as English and German, while many languages are severely underrepresented. This mirrors broader concerns about fairness and robustness in multilingual speaker recognition~\cite{peri2022train}. Our results indicate that simple language subspace compensation improves average performance and generalizes to unseen languages. Because NAP is estimated globally across all languages, it may under-compensate for very low-resource languages that behave differently from the dominant ones, which motivates future work on family-specific or hierarchical subspaces. A more fine-grained analysis per language family and resource level is needed to fully assess fairness.

\section{Conclusion}

We presented our TidyVoice2026 submission on cross-lingual verification. Using a SimAM ResNet34 pretrained on VoxCeleb and VoxBlink2 and fine-tuned on TidyVoice~\cite{qin2022simple,snyder2018x,farhadipour2026tidyvoice,farhadipour2026tidyvoicepaper}, we showed NAP-based language normalization with AS-Norm reduces EER from 2.70 to 2.18 and yields a Codabench score of 8.40. More complex alternatives, including ECAPA TDNN encoders, adversarial language adaptation, LDA and PLDA back-ends, and WavLM-based models, did not surpass the post hoc pipeline on this benchmark~\cite{desplanques2020ecapa,chen2022wavlm,lin2019semi,rohdin2019speaker}. These findings reinforce the value of classical nuisance subspace modeling in speaker verification, particularly under cross-language mismatch where language labels are not available at test time~\cite{dehak2010front,misra2018modelling}. Future work will investigate whether NAP-style language compensation can be combined with domain adaptation techniques, and how self-supervised encoders such as WavLM can be adapted to multilingual, imbalanced corpora like TidyVoice without sacrificing speaker discriminability. Because the proposed language-normalization step operates only on embeddings, it can be reused across TidyVoice-style evaluations and integrated into speaker verification pipelines without modifying training or requiring language labels at inference time.

\section{Generative AI Use Disclosure}

LLMs were used exclusively for language editing, including rephrasing and grammatical refinement, to improve clarity and readability. The LLMs were not involved in the development of ideas, methodology design, experimental procedures, data analysis, or interpretation of results. All scientific content was developed and verified by the author/s.

\bibliographystyle{IEEEtran}
\bibliography{mybib}

\end{document}